\documentclass[11pt,a4paper]{article}
\usepackage[hyperref]{acl2019}
\usepackage{times}
\usepackage{latexsym}
\usepackage{graphicx}
\usepackage{url}
\usepackage{amsmath,amsthm,amssymb}
\usepackage[FIGTOPCAP]{subfigure}
\usepackage{tabularx}
\usepackage{multirow}
\usepackage{makecell}
\usepackage{xspace}
\usepackage{todonotes}
\usepackage{mdwlist}
\usepackage{paralist}

\aclfinalcopy 

\newcommand{\bh}{\mathbf{h}}
\newcommand{\bq}{\mathbf{q}}
\newcommand{\bc}{\mathbf{c}}
\newcommand{\bQ}{\mathbf{Q}}
\newcommand{\bC}{\mathbf{C}}
\newcommand{\mR}{\mathbb{R}}

\newcommand{\bO}{\mathbf{O}}
\newcommand{\bW}{\mathbf{W}}
\newcommand{\be}{\mathbf{e}}
\newcommand{\bE}{\mathbf{E}}
\newcommand{\bb}{\mathbf{b}}
\newcommand{\bm}{\mathbf{m}}
\newcommand{\bM}{\mathbf{M}}

\newcommand{\hotpotqa}{HotpotQA\xspace}
\newcommand{\method}{DFGN\xspace}

\title{Dynamically Fused Graph Network for Multi-hop Reasoning}
\aclfinalcopy

\author{
  Yunxuan Xiao$^{1\dagger}$ \quad Yanru Qu$^{1\dagger}$ \quad Lin Qiu$^{1\dagger}$ \\
  \textbf{Hao Zhou$^{2}$ \quad Lei Li$^{2}$} \quad
  \textbf{Weinan Zhang$^{1}$ \quad Yong Yu$^{1}$} \\
  {$^1$~Shanghai Jiao Tong University  \quad $^2$~ByteDance AI Lab, China}\\
  {\tt \{xiaoyunxuan, wnzhang\}@sjtu.edu.cn}\\
  {\tt \{lqiu, kevinqu, yyu\}@apex.sjtu.edu.cn}\\
  {\tt \{zhouhao.nlp, lilei.02\}@bytedance.com} \\
}

\date{}

\begin{document}
\maketitle
\renewcommand{\thefootnote}{\fnsymbol{footnote}}
\footnotetext[2]{These authors contributed equally. The order of authorship is decided through dice rolling. Work done while Lin Qiu was a research intern in ByteDance AI Lab.}
\renewcommand{\thefootnote}{\arabic{footnote}}
\begin{abstract}
Text-based question answering (TBQA) has been studied extensively in recent years.
Most existing approaches focus on finding the answer to a question within a single paragraph. 
However, many difficult questions require multiple supporting evidence from scattered text across two or more documents. 
In this paper, we propose the Dynamically Fused Graph Network~(\method), 
a novel method to answer those questions requiring multiple scattered evidence and reasoning over them. 
Inspired by human's step-by-step reasoning behavior, 
\method includes a dynamic fusion layer
that starts from the entities mentioned in the given query, 
explores along the entity graph dynamically built from the text, 
and gradually finds relevant supporting entities from the given documents. 
We evaluate \method on \hotpotqa, a public TBQA dataset requiring multi-hop reasoning. 
\method achieves competitive results on the public board. 
Furthermore, our analysis shows \method could produce interpretable reasoning chains. 
\end{abstract}

\section{Introduction}
\label{sec:intro}
\begin{figure}[t]
\begin{center}
\includegraphics[width=\columnwidth]{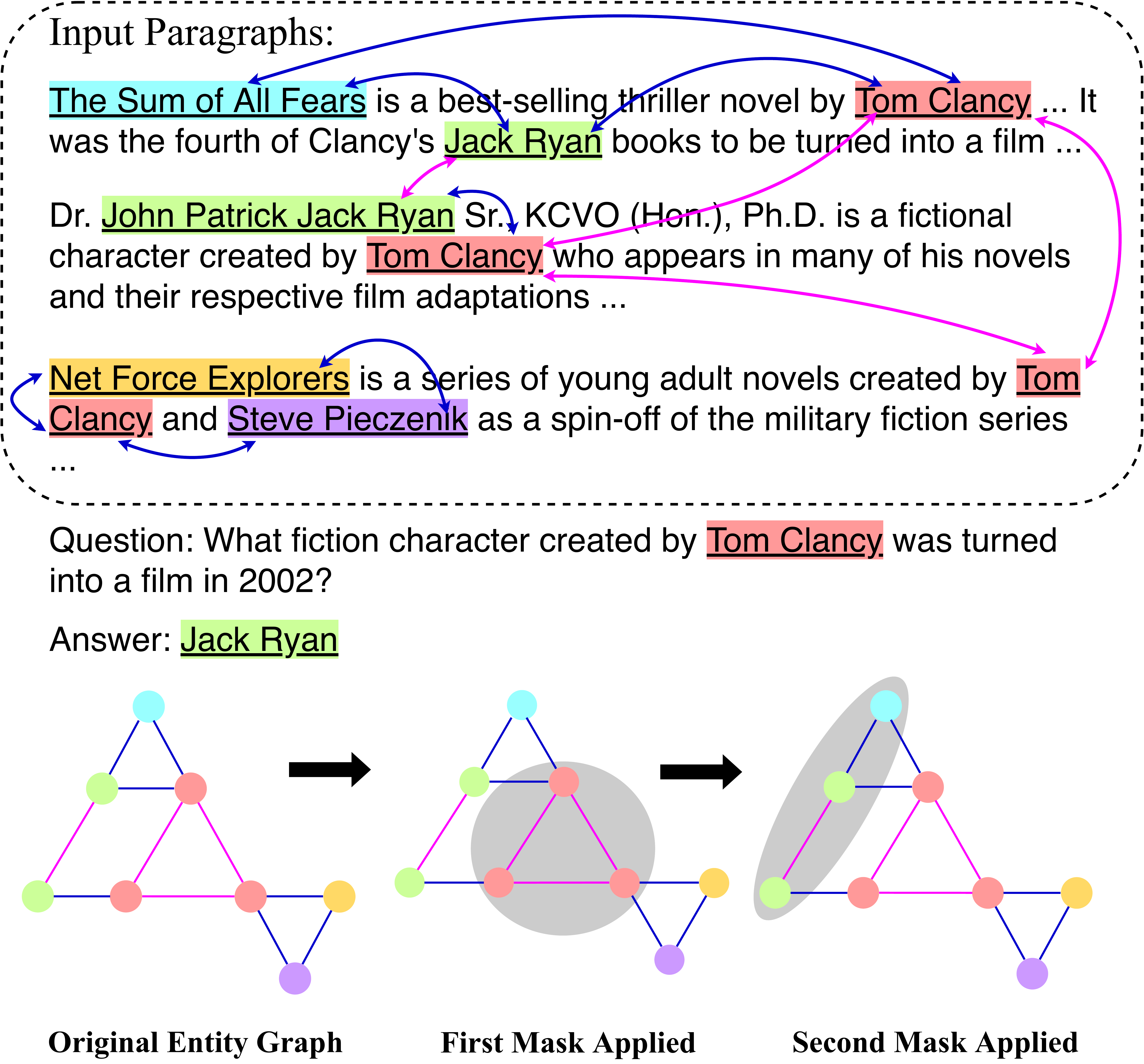}
\end{center}
\caption{Example of multi-hop text-based QA. One question and three document paragraphs are given. 
Our proposed \method conducts multi-step reasoning over the facts by constructing an entity graph from multiple paragraphs,  
predicting a dynamic mask to select a sub-graph, propagating information along the graph, and finally transfer the information from the graph back to the text in order to localize the answer. 
Nodes are entity occurrences, with the color denoting the underlying entity. Edges are constructed from co-occurrences. 
The gray circles are selected by \method in each step.
}
\label{fig:case}
\end{figure}

Question answering (QA) has been a popular topic in natural language processing. 
QA provides a quantifiable way to evaluate an NLP system's capability on language understanding and reasoning~\cite{hermann2015teaching,rajpurkar2016squad,rajpurkar2018know}.
Most previous work focus on finding evidence and answers from a single paragraph~\cite{seo2016bidirectional,liu2017stochastic,wang2017gated}.
It rarely tests deep reasoning capabilities of the underlying model. 
In fact, \newcite{min2018efficient} observe that most questions in existing QA benchmarks can be answered by retrieving a small set of sentences without reasoning. 
To address this issue, there are several recently proposed QA datasets particularly designed to evaluate a system's 
multi-hop reasoning capabilities, including WikiHop~\cite{welbl2018constructing}, ComplexWebQuestions~\cite{talmor2018web}, and HotpotQA~\cite{yang2018hotpotqa}.

In this paper, we study the problem of multi-hop text-based QA, 
which requires multi-hop reasoning among evidence scattered around multiple raw documents. 
In particular, a query utterance and a set of accompanying documents are given, but not all of them are relevant. 
The answer can only be obtained by selecting two or more evidence from the documents and inferring among them (see Figure~\ref{fig:case} for an example). 
This setup is versatile and does not rely on any additional predefined knowledge base. 
Therefore the models are expected to generalize well and to answer questions in open domains. 

There are two main challenges to answer questions of this kind. 
Firstly, since not every document contain relevant information, 
multi-hop text-based QA requires filtering out noises from multiple paragraphs and extracting useful information.
To address this, recent studies propose to build entity graphs from input paragraphs and apply graph neural networks (GNNs) to aggregate the information through entity graphs \cite{dhingra2018neural,de2018question,song2018exploring}.
However, all of the existing work apply GNNs based on a static global entity graph of each QA pair, which can be considered as performing implicit reasoning.
Instead of them, we argue that the query-guided multi-hop reasoning should be explicitly performed on a dynamic local entity graph tailored according to the query. 

Secondly, previous work on multi-hop QA (e.g. WikiHop) usually aggregates document information to an entity graph, and answers are then directly selected on entities of the entity graph.
However, in a more realistic setting, the answers may even not reside in entities of the extracted entity graph.
Thus, existing approaches can hardly be directly applied to open-domain multi-hop QA tasks like \hotpotqa.


In this paper, we propose Dynamically Fused Graph Network (\method), a novel method to address the aforementioned concerns for multi-hop text-based QA.
For the first challenge, \method constructs a \emph{dynamic entity graph} based on entity mentions in the query and documents. 
This process iterates in multiple rounds to achieve multi-hop reasoning. 
In each round, \method generates and reasons on a dynamic graph, where irrelevant entities are masked out while only reasoning sources are preserved, via a mask prediction module. 
Figure~\ref{fig:case} shows how DFGN works on a multi-hop text-based QA example in HotpotQA.
The mask prediction module is learned in an end-to-end fashion, alleviating the error propagation problem.

To solve the second challenge, we propose a \emph{fusion process} in \method to solve the unrestricted QA challenge.
We not only aggregate information from documents to the entity graph (doc2graph), but also propagate the information of the entity graph back to document representations (graph2doc).
The fusion process is iteratively performed at each hop through the document tokens and entities, and the final resulting answer is then obtained from document tokens.
The \emph{fusion process} of doc2graph and graph2doc along with the \emph{dynamic entity graph} jointly improve the interaction between the information of documents and the entity graph, leading to a less noisy entity graph and thus more accurate answers.

As one merit, \method's predicted masks implicitly induce reasoning chains, which can explain the reasoning results. 
Since the ground truth reasoning chain is very hard to define and label for open-domain corpus, we propose a feasible way to weakly supervise the mask learning. 
We propose a new metric to evaluate the quality of predicted reasoning chains and constructed entity graphs.

Our contributions are summarized as follows:
\begin{itemize*}
    \item We propose \method, a novel method for the multi-hop text-based QA problem.
    \item We provide a way to explain and evaluate the reasoning chains via interpreting the entity graph masks predicted by \method. The mask prediction module is additionally weakly trained.
    \item We provide an experimental study on a public dataset~(\hotpotqa) to demonstrate that our proposed \method is competitive against state-of-the-art unpublished work. 
\end{itemize*}

\section{Related work}
\label{sec:related}
\begin{figure*}[t]
\begin{center}
\includegraphics[width=2.05\columnwidth]{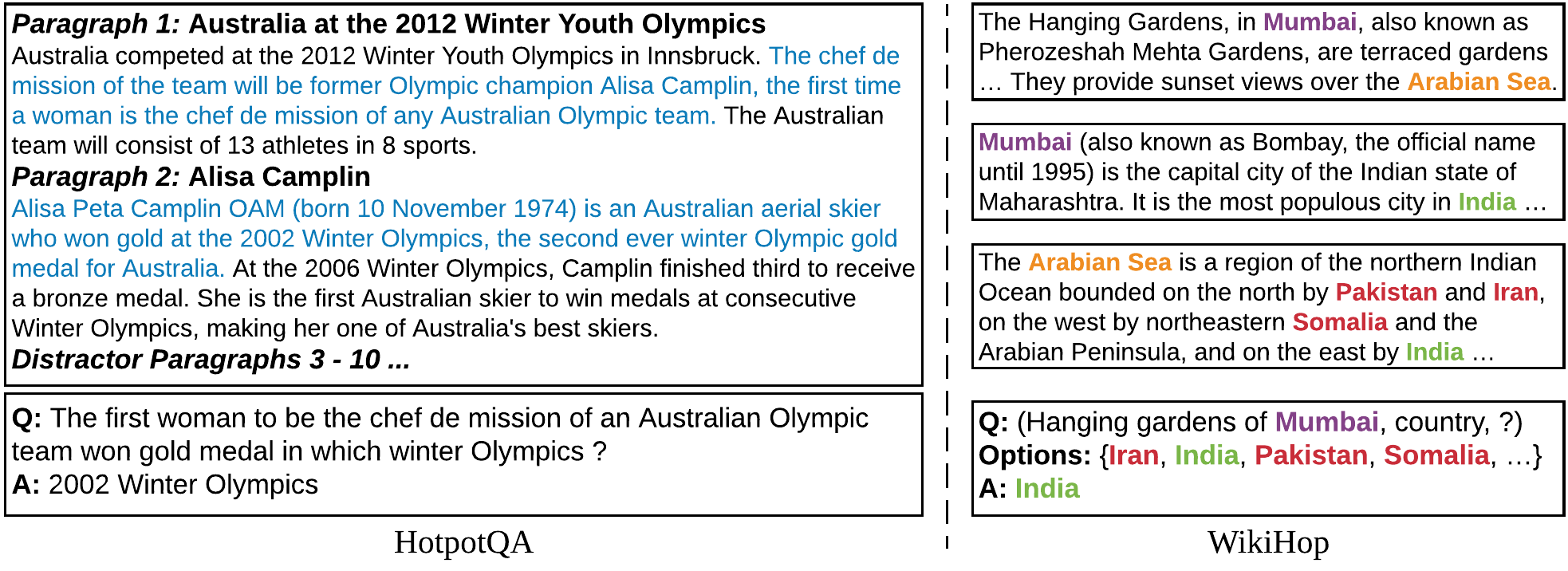}
\end{center}
\caption{Comparison between HotpotQA (left) and WikiHop (right). In HotpotQA, the questions are proposed by crowd workers and the blue words in paragraphs are labeled supporting facts corresponding to the question. In WikiHop, the questions and answers are formed with relations and entities in the underlying KB respectively, thus the questions are inherently restricted by the KB schema. The colored words and phrases are entities in the KB.}
\label{fig:dataset}
\end{figure*}

\paragraph{Text-based Question Answering}
Depending on whether the supporting information is structured or not, 
QA tasks can be categorized into knowledge-based (KBQA), text-based (TBQA), mixed, and others. 
In KBQA, the supporting information is from structured knowledge bases (KBs), while the queries can be either structure or natural language utterances. 
For example, SimpleQuestions is one large scale dataset of this kind~\cite{bordes2015large}.
In contrast, TBQA's supporting information is raw text, and hence the query is also text.
SQuAD~\cite{rajpurkar2016squad} and \hotpotqa~\cite{yang2018hotpotqa} are two such datasets. 
There are also mixed QA tasks which combine both text and KBs, e.g. WikiHop~\cite{welbl2018constructing} and ComplexWebQuestions~\cite{talmor2018web}.
In this paper, we focus on TBQA, since TBQA tests a system's end-to-end capability of extracting relevant facts from raw language and reasoning about them. 


Depending on the complexity in underlying reasoning, QA problems can be categorized into single-hop and multi-hop ones. 
Single-hop QA only requires one fact extracted from the underlying information, no matter structured or unstructured, 
e.g. ``which city is the capital of California''.
The SQuAD dataset belongs to this type~\cite{rajpurkar2016squad}.
On the contrary, multi-hop QA requires identifying multiple related facts 
and reasoning about them, e.g. ``what is the capital city of the largest state in the U.S.''.
Example tasks and benchmarks of this kind include WikiHop, ComplexWebQuestions, and \hotpotqa.
Many IR techniques can be applied to answer single-hop questions~\cite{rajpurkar2016squad}. 
However, these IR techniques are hardly introduced in multi-hop QA, since a single fact can only partially match a question.


Note that existing multi-hop QA datasets WikiHop and ComplexWebQuestions, are constructed using existing KBs and constrained by the schema of the KBs they use. For example, the answers are limited in entities in WikiHop rather than formed by free texts in \hotpotqa (see Figure~\ref{fig:dataset} for an example). In this work, we focus on multi-hop text-based QA, so we only evaluate on \hotpotqa. 


\paragraph{Multi-hop Reasoning for QA}
Popular GNN frameworks, e.g. graph convolution network \cite{kipf2016semi}, graph attention network \cite{velivckovic2017graph}, and graph recurrent network \cite{song2018graph}, have been previously studied and show promising results in QA tasks requiring reasoning~\cite{dhingra2018neural,de2018question,song2018exploring}.

Coref-GRN extracts and aggregates entity information in different references from scattered paragraphs~\cite{dhingra2018neural}. 
Coref-GRN utilizes co-reference resolution to detect different mentions of the same entity. 
These mentions are combined with a graph recurrent neural network (GRN)~\cite{song2018graph} to produce aggregated entity representations. 
MHQA-GRN \cite{song2018exploring} follows Coref-GRN and refines the graph construction procedure with more connections: sliding-window, same entity, and co-reference, which shows further improvements. 
Entity-GCN \cite{de2018question} proposes to distinguish different relations in the graphs through a relational graph convolutional neural network (GCN)~\cite{kipf2016semi}.
Coref-GRN, MHQA-GRN and Entity-GCN explore the graph construction problem in answering real-world questions. 
However, it is yet to investigate how to effectively reason about the constructed graphs, which is the main problem studied in this work.

Another group of sequential models deals with multi-hop reasoning following Memory Networks \cite{sukhbaatar2015end}. 
Such models construct representations for queries and memory cells for contexts, then make interactions between them in a multi-hop manner. \newcite{munkhdalai2016reasoning} and \newcite{onishi2016did} incorporate a hypothesis testing loop to update the query representation at each reasoning step and select the best answer among the candidate entities at the last step. 
IRNet \cite{zhou2018interpretable} generates a subject state and a relation state at each step, computing the similarity score between all the entities and relations given by the dataset KB. 
The ones with the highest score at each time step are linked together to form an interpretable reasoning chain. 
However, these models perform reasoning on simple synthetic datasets with a limited number of entities and relations, which are quite different with large-scale QA dataset with complex questions.
Also, the supervision of entity-level reasoning chains in synthetic datasets can be easily given following some patterns while they are not available in \hotpotqa. 

\section{Dynamically Fused Graph Network}
\label{sec:method}
\begin{figure}[tb]
\centering
\includegraphics[width=0.7\columnwidth]{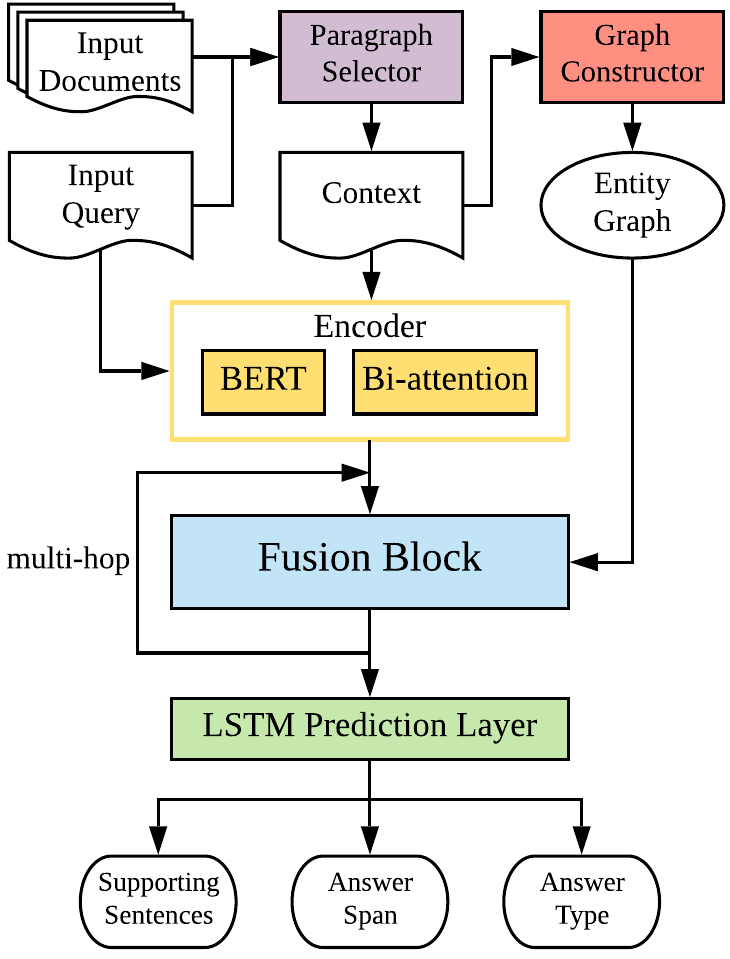}
\caption{Overview of \method.}
\label{fig:dfgn-system}
\end{figure}

We describe dynamically fused graph network~(\method) in this section.
Our intuition is drawn from the human reasoning process for QA. 
One starts from an entity of interest in the query, 
focuses on the words surrounding the start entities, 
connects to some related entity either found in the neighborhood 
or linked by the same surface mention, 
repeats the step to form a reasoning chain, 
and lands on some entity or snippets likely to be the answer. 
To mimic human reasoning behavior, we develop five components 
in our proposed QA system~(Fig.~\ref{fig:dfgn-system}): a paragraph selection sub-network, a module for entity graph construction, an encoding layer, 
a fusion block for multi-hop reasoning, and a final prediction layer.


\subsection{Paragraph Selection} 
For each question, we assume that $N_p$ paragraphs are given
(e.g. $N_p=10$ in \hotpotqa). 
Since not every piece of text is relevant to the question, 
we train a sub-network to select relevant paragraphs. 
The sub-network is based on a pre-trained BERT model~\cite{devlin2018bert}
followed by a sentence classification layer with sigmoid prediction.
The selector network takes a query $Q$ and a paragraph as input and outputs a relevance score between 0 and 1. 
Training labels are constructed by assigning 1's to the paragraphs with at least one supporting sentence for each Q\&A pair.
During inference, paragraphs with predicted scores greater than $\eta$ ($=0.1$ in experiments) are selected and concatenated together as the context $C$. 
$\eta$ is properly chosen to ensure the selector reaches a significantly high recall of relevant paragraphs. 
$Q$ and $C$ are further processed by upper layers. 

\subsection{Constructing Entity Graph}
We do not assume a global knowledge base. 
Instead, we use the Stanford corenlp toolkit~\cite{manning2014stanford} to recognize named entities from the context $C$. 
The number of extracted entities is denoted as $N$.
The entity graph is constructed with the entities as nodes and edges built as follows. 
The edges are added
\begin{inparaenum}
\item for every pair of entities appear in the same sentence in $C$ (sentence-level links);
\item for every pair of entities with the same mention text in $C$ (context-level links);
and 
\item between a central entity node and other entities within the same paragraph (paragraph-level links).
\end{inparaenum}
The central entities are extracted from the title sentence for each paragraph.
Notice the context-level links ensures that entities across multiple documents are connected in a certain way. 
We do not apply co-reference resolution for pronouns because it introduces both additional useful and erroneous links. 




\subsection{Encoding Query and Context}
We concatenate the query $Q$ with the context $C$ 
and pass the resulting sequence to a pre-trained BERT model to obtain representations
$\bQ = [\bq_1, \dots, \bq_{L}]\in \mR^{L \times d_1}$ and $\bC^\top = [\bc_1, \dots, \bc_{M}] \in \mR^{M \times d_1}$, 
where $L$,$M$ are lengths of query and context,  and $d_1$ is the size of BERT hidden states. 
In experiments, we find concatenating queries and contexts performs better than passing them separately to BERT.

The representations are further passed through a  bi-attention layer~\cite{seo2016bidirectional} to enhance cross interactions between the query and the context. 
In practice, we find adding the bi-attention layer achieves better performance than the BERT encoding only. 
The output representation are $\bQ_{0} \in \mR^{L \times d_2}$ and $\bC_{0} \in \mR^{M \times d_2}$, where $d_2$ is the output embedding size.



\subsection{Reasoning with the Fusion Block}
\begin{figure}[tb]
\centering
\includegraphics[width=\columnwidth]{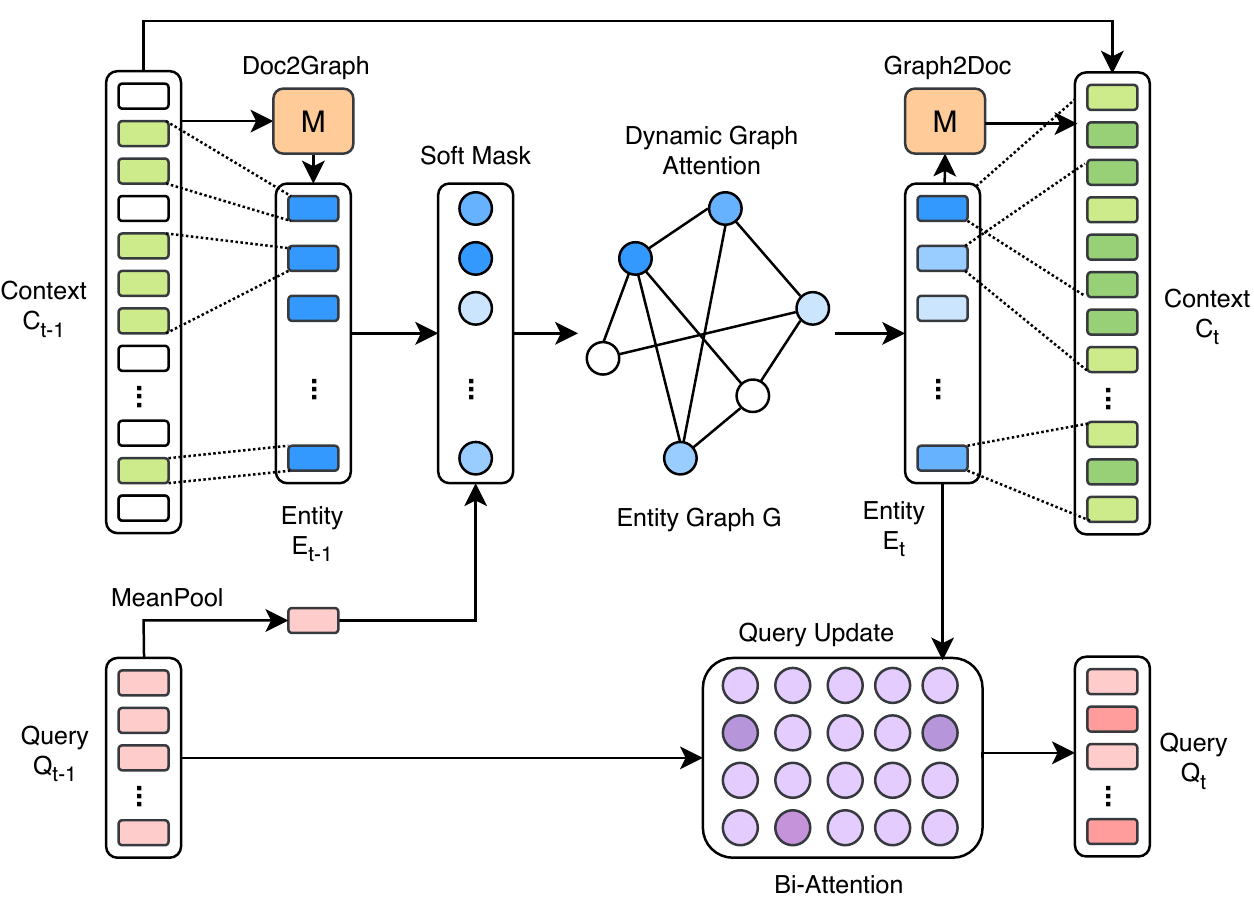}
\caption{Reasoning with the fusion block in \method}
\label{fig:gnn}
\end{figure}

With the embeddings calculated for the query $Q$ and context $C$, 
the remaining challenge is how to identify supporting entities and the text span of potential answers. 
We propose a fusion block to mimic human's one-step reasoning behavior -- starting from $\bQ_0$ and $\bC_{0}$ and finding one-step supporting entities. 
A fusion block achieves the following:
\begin{inparaenum}
    \item passing information from tokens to entities by computing entity embeddings from tokens (Doc2Graph flow);
    \item propagating information on entity graph; and
    \item passing information from entity graph to document tokens since the final prediction is on tokens (Graph2Doc flow). 
\end{inparaenum}
Fig.~\ref{fig:gnn} depicts the inside structure of the fusion block in \method.

\paragraph{Document to Graph Flow.}
Since each entity is recognized via the NER tool, the text spans associated with the entities are utilized to compute entity embeddings (Doc2Graph). 
To this end, we construct a binary matrix $\bM$, where $\bM_{i,j}$ is 1 if 
$i$-th token in the context is within the span of the $j$-th entity. 
$\bM$ is used to select the text span associated with an entity. 
The token embeddings calculated from the above section (which is a matrix containing only selected columns of $\bC_{t-1}$) is passed into a mean-max pooling to calculate entity embeddings $\bE_{t-1} = [\be_{t-1,1}, \dots, \be_{t-1, N}]$. 
$\bE_{t-1}$ will be of size $2d_2 \times N$, where $N$ is the number of entities, and each of the $2d_2$ dimensions will produce both mean-pooling and max-pooling results. 
This module is denoted as Tok2Ent.


\paragraph{Dynamic Graph Attention.}
After obtaining entity embeddings from the input context $\bC_{t-1}$, we apply a graph neural network to propagate node information to their neighbors. 
We propose a dynamic graph attention mechanism to mimic human's step-by-step exploring and reasoning behavior. 
In each reasoning step, we assume every node has some information to disseminate to neighbors. 
The more relevant to the query, the neighbor nodes receive more information from nearby. 


We first identify nodes relevant to the query by creating a soft mask on entities. 
It serves as an information gatekeeper, i.e. only those entity nodes pertaining to the query are allowed to disseminate information. 
We use an attention network between the query embeddings and the entity embeddings to predict a soft mask $\bm_t$, which aims to signify the start entities in the $t$-th reasoning step:
\begin{align}
\tilde{\bq}^{(t-1)} & = \text{MeanPooling}(\bQ^{(t-1)}) \\
\gamma^{(t)}_i &= \Tilde{\mathbf{q}}^{(t-1)} \mathbf{V}^{(t)} \mathbf{e}^{(t-1)}_i / \sqrt{d_2} \\
\bm^{(t)} & = \sigma ([\gamma^{(t)}_1,\cdots,\gamma^{(t)}_N]) \label{eq:mask} \\
\tilde{\bE}^{(t-1)} & = [m^{(t)}_1 \be^{(t-1)}_1, \dots, m^{(t)}_{N} \be^{(t-1)}_{N}] \label{eq:masked}
\end{align}
where $\mathbf{V}_t$ is a linear projection matrix,  and $\sigma$ is the sigmoid function.
By multiplying the soft mask and the initial entity embeddings, the desired start entities will be encouraged and others will be penalized. As a result, this step of information propagation is restricted to a dynamic sub-part of the entity graph.

The next step is to disseminate information across the dynamic sub-graph.
Inspired by GAT~\cite{velivckovic2017graph}, we compute attention score $\alpha$ between two entities by:
\begin{align}
\bh^{(t)}_{i} & = \mathbf{U}_t \tilde{\be}^{(t-1)}_{i} + \bb_t \\
\beta^{(t)}_{i,j} & = \text{LeakyReLU}({\bW^\top_{t}} [\bh^{(t)}_{i}, \bh^{(t)}_{j}]) \\
\alpha^{(t)}_{i,j} & = \frac{\exp(\beta^{(t)}_{i,j})}{\sum_k \exp(\beta^{(t)}_{i,k})} \label{eq:attn}
\end{align}
where $\mathbf{U}_t \in \mR^{d_2 \times 2d_2}$, $\bW_{t} \in \mR^{2d_2}$ are linear projection parameters.
Here the $i$-th row of $\alpha$ represents the proportion of information that will be assigned to the neighbors of entity $i$. 

Note that the information flow in our model is different from most previous GATs. In dynamic graph attention, each node sums over its column, which forms a new entity state containing the total information it received from the neighbors:
\begin{align}
\be^{(t)}_{i} & = \text{ReLU}(\sum_{j \in B_i} \alpha^{(t)}_{j,i} \bh^{(t)}_{j}) \label{eq:alloc}
\end{align}
where $B_i$ is the set of neighbors of entity $i$. Then we obtain the updated entity embeddings $\bE^{(t)} = [\be^{(t)}_{1}, \dots, \be^{(t)}_{N}]$.



\paragraph{Updating Query.}
A reasoning chain contains multiple steps, and the newly visited entities by one step will be the start entities of the next step.
In order to predict the expected start entities for the next step,
we introduce a query update mechanism, where the query embeddings are updated by the entity embeddings of the current step. In our implementation, we utilize a bi-attention network \cite{seo2016bidirectional} to update the query embeddings:
\begin{align}
\bQ^{(t)} = \text{Bi-Attention}(\bQ^{(t-1)}, \bE^{(t)})
\end{align}

\paragraph{Graph to Document Flow.}
Using Tok2Ent and dynamic graph attention, we realize a reasoning step at the entity level. 
However, the unrestricted answer still cannot be backtraced. 
To address this, we develop a Graph2Doc module to keep information flowing from entity back to tokens in the context. 
Therefore the text span pertaining to the answers can be localized in the context. 

Using the same binary matrix $\bM$ as described above, 
the previous token embeddings in $\bC_{t-1}$ are concatenated with the associated entity embedding corresponding to the token. 
Each row in $\bM$ corresponds to one token, therefore we use it to select one entity's embedding from $\bE_t$ if the token participates in the entity's mention. 
This information is further processed with a LSTM layer \cite{hochreiter1997long} to produce the next-level context representation:
\begin{align}
\bC^{(t)} = \text{LSTM}([\bC^{(t-1)}, \bM \bE^{(t)\top}])
\end{align}
where $;$ refers to concatenation and $\bC^{(t)} \in \mR^{M \times d_2}$ serves as the input of the next fusion block. At this time, the reasoning information of current sub-graph has been propagated onto the whole context.

\subsection{Prediction}
We follow the same structure of prediction layers as \cite{yang2018hotpotqa}. 
The framework has four output dimensions, including
\begin{inparaenum}
    \item supporting sentences, 
    \item the start position of the answer,
    \item the end position of the answer, and
    \item the answer type.
\end{inparaenum}
We use a cascade structure to solve the output dependency, where four isomorphic LSTMs $\mathcal{F}_i$ are stacked layer by layer. The context representation of the last fusion block is sent to the first LSTM $\mathcal{F}_0$. Each $\mathcal{F}_i$ outputs a logit $\bO \in \mR^{M \times d_2}$ and computes a cross entropy loss over these logits. 
\begin{align}
\bO_{sup} & = \mathcal{F}_0(\bC^{(t)}) \\
\bO_{start} & = \mathcal{F}_1([\bC^{(t)}, \bO_{sup}]) \\
\bO_{end} & = \mathcal{F}_2([\bC^{(t)}, \bO_{sup}, \bO_{start}]) \\
\bO_{type} & = \mathcal{F}_3([\bC^{(t)}, \bO_{sup}, \bO_{end}])
\end{align}

We jointly optimize these four cross entropy losses. Each loss term is weighted by a coefficient.
\begin{align}
    \mathcal{L} = \mathcal{L}_{start} + \mathcal{L}_{end} + \lambda_s \mathcal{L}_{sup} + \lambda_t \mathcal{L}_{type}
\end{align}

\paragraph{Weak Supervision.}
In addition, we introduce a weakly supervised signal to induce the soft masks at each fusion block to match the heuristic masks. 
For each training case, the heuristic masks contain a start mask detected from the query, and additional BFS masks obtained by applying breadth-first search (BFS) on the adjacent matrices give the start mask. 
A binary cross entropy loss between the predicted soft masks and the heuristics is then added to the objective.
We skip those cases whose start masks cannot be detected from the queries.

\section{Experiments}
\label{sec:exp}
\begin{table*}[t]
    \centering 
	\begin{tabular}{ccccccc}
		\Xhline{1pt}
		\multirow{2}{*}{Model} & \multicolumn{2}{c}{Answer} 	& \multicolumn{2}{c}{Sup Fact} & \multicolumn{2}{c}{Joint} \\ \cline{2-7}
		& EM & F1 & EM & F1 & EM & F1\\ \hline
		\multicolumn{1}{l}{Baseline Model} & 45.60 & 59.02 & 20.32 & 64.49 & 10.83 & 40.16 \\
		\multicolumn{1}{l}{GRN$^*$} & 52.92 & 66.71 & 52.37 & 84.11 & 31.77 & 58.47 \\
		\multicolumn{1}{l}{DFGN(Ours)} & 55.17 & 68.49 & 49.85 & 81.06 & 31.87 & 58.23 \\
		\multicolumn{1}{l}{QFE$^*$} & 53.86 & 68.06 & \textbf{57.75} & \textbf{84.49} & \textbf{34.63} & 59.61\\ \hline
		\multicolumn{1}{l}{DFGN(Ours)$\dagger$} & \textbf{56.31} & \textbf{69.69} & 51.50 & 81.62 & 33.62 & \textbf{59.82}\\
		\Xhline{1pt}
	\end{tabular}
	\caption{Performance comparison on the private test set of HotpotQA in the distractor setting. Our DFGN is the second best result on the leaderboard before submission (on March 1st). The baseline model is from \newcite{yang2018hotpotqa} and the results with $^*$ is unpublished. DFGN(Ours)$\dagger$ refers to the same model with a revised entity graph, whose entities are recognized by a BERT NER model. Note that the result of DFGN(Ours)$\dagger$ is submitted to the leaderboard during the review process of our paper.} 
	\label{table:main}
\end{table*}
\begin{table}[t]
    \centering 
	\begin{tabular}{lcc}
	    Setting & EM & F1\\ \Xhline{1pt}
	    DFGN (2-layer) & 55.42 & 69.23 \\
	    - BFS Supervision & 54.48 & 68.15 \\
	    - Entity Mask & 54.64 & 68.25 \\
	    - Query Update & 54.44 & 67.98 \\
	    - E2T Process & 53.91 & 67.45 \\
	    - 1 Fusion Block & 54.14 & 67.70 \\
	    - 2 Fusion Blocks & 53.44 & 67.11 \\
	    - 2 Fusion Blocks \& Bi-attn & 50.03 & 62.83 \\
		\Xhline{1pt}
		gold paragraphs only & 55.67 & 69.15 \\
		supporting facts only & 57.57 & 71.67 \\
		\Xhline{1pt}
	\end{tabular}
	\caption{Ablation study of question answering performances in the development set of HotpotQA in the distractor setting. We use a DFGN with 2-layer fusion blocks as the origin model. The upper part is the model ablation results and the lower part is the dataset ablation results.}
	\label{table:ablation}
\end{table}

We evaluate our Dynamically Fused Graph Network (DFGN) on HotpotQA \cite{yang2018hotpotqa} in the distractor setting. For the full wiki setting where the entire Wikipedia articles are given as input, we consider the bottleneck is about information retrieval, thus we do not include the full wiki setting in our experiments.

\subsection{Implementation Details}

In paragraph selection stage, we use the uncased version of BERT Tokenizer~\cite{devlin2018bert} to tokenize all passages and questions. 
The encoding vectors of sentence pairs are generated from a pre-trained BERT model~\cite{devlin2018bert}.
We set a relatively low threshold during selection to keep a high recall (97$\%$) and a reasonable precision (69$\%$) on supporting facts.

In graph construction stage, we use a pre-trained NER model from Stanford CoreNLP Toolkits\footnote{\url{https://nlp.stanford.edu/software/CRF-NER.shtml}} \cite{manning2014stanford} to extract named entities. The maximum number of entities in a graph is set to be 40. Each entity node in the entity graphs has an average degree of 3.52.

In the encoding stage, we also use a pre-trained BERT model as the encoder, thus $d_1$ is 768. All the hidden state dimensions $d_2$ are set to 300. We set the dropout rate for all hidden units of LSTM and dynamic graph attention to 0.3 and 0.5 respectively. For optimization, we use Adam Optimizer \cite{kingma2014adam} with an initial learning rate of 1e$^{-4}$.


\subsection{Main Results}
We first present a comparison between baseline models and our DFGN\footnote{Our code is available in \url{https://github.com/woshiyyya/DFGN-pytorch}.}. Table \ref{table:main} shows the performance of different models in the private test set of HotpotQA. From the table we can see that our model achieves the second best result on the leaderboard now\footnote{The leaderboard can be found on \url{https://hotpotqa.github.io}} (on March 1st). Besides, the answer performance and the joint performance of our model are competitive against state-of-the-art unpublished models.
We also include the result of our model with a revised entity graph whose entities are recognized by a BERT NER model \cite{devlin2018bert}.
We fine-tune the pre-trained BERT model on the dataset of the CoNLL'03 NER shared task \cite{sang2003introduction} and use it to extract named entities from the input paragraphs.
The results show that our model achieves a 1.5\% gain in the joint F1-score with the entity graph built from a better entity recognizer.

To evaluate the performance of different components in our DFGN, we perform ablation study on both model components and dataset segments. Here we follow the experiment setting in \newcite{yang2018hotpotqa} to perform the dataset ablation study, where we only use golden paragraphs or supporting facts as the input context. The ablation results of QA performances in the development set of HotpotQA are shown in Table \ref{table:ablation}. From the table we can see that each of our model components can provide from 1\% to 2\% relative gain over the QA performance. 
Particularly, using a 1-layer fusion block leads to an obvious performance loss, which implies the significance of performing multi-hop reasoning in HotpotQA. Besides, the dataset ablation results show that our model is not very sensitive to the noisy paragraphs comparing with the baseline model which can achieve a more than 5\% performance gain in the ``gold paragraphs only'' and ``supporting facts only'' settings. \cite{yang2018hotpotqa}.

\subsection{Evaluation on Graph Construction and Reasoning Chains}
The chain of reasoning is a directed path on the entity graph, so high-quality entity graphs are the basis of good reasoning. 
Since the limited accuracy of NER model and the incompleteness of our graph construction, 31.3\% of the cases in the development set are unable to perform a complete reasoning process, where at least one supporting sentence is not reachable through the entity graph, i.e. no entity is recognized by NER model in this sentence. 
We name such cases as ``missing supporting entity'', and the ratio of such cases can evaluate the quality of graph construction.
We focus on the rest 68.7\% good cases in the following analysis.

In the following, we first give several definitions before presenting ESP (Entity-level Support) scores.

\paragraph{Path} A path is a sequence of entities visited by the fusion blocks, denoting as $P = [e_{p_1}, \dots, e_{p_{t+1}}]$ (suppose $t$-layer fusion blocks).

\paragraph{Path Score} The score of a path is acquired by multiplying corresponding soft masks and attention scores along the path, i.e. $score(P) = \prod_{i=1}^{t} m^{(i)}_{p_i} \alpha^{(i)}_{p_i,p_{i+1}}$ (Eq.~\eqref{eq:mask}, \eqref{eq:attn}).

\paragraph{Hit} Given a path and a supporting sentence, if at least one entity of the supporting sentence is visited by the path, we call this supporting sentence is hit\footnote{A supporting sentence may contain irrelevant information, thus we do not have to visit all entities in a supporting sentence. 
Besides, due to the fusion mechanism of \method, the entity information will be propagated to the whole sentence. Therefore, we define a ``hit'' occurs when at least one entity of the supporting sentence is visited.}.

Given a case with $m$ supporting sentences, we select the top-$k$ paths with the highest scores as the predicted reasoning chains. For each supporting sentence, we use the $k$ paths to calculate how many supporting sentences are hit. 

In the following, we introduce two metrics to evaluate the quality of multi-hop reasoning through entity-level supporting (ESP) scores.

\paragraph{ESP EM (Exact Match)} 
For a case with $m$ supporting sentences, if all the $m$ sentences are hit, we call this case exact match. The ESP EM score is the ratio of exactly matched cases. 
\paragraph{ESP Recall} 
For a case with $m$ supporting sentences and $h$ of them are hit, this case has a recall score of $h/m$. The averaged recall of the whole dataset is the ESP Recall.

\begin{table}[t]
    \centering
    \begin{tabular}{c|p{0.7cm}<{\centering}p{0.7cm}<{\centering}p{0.7cm}<{\centering}p{0.7cm}<{\centering}} \Xhline{1pt}
        k & 1 & 2 & 5 & 10 \\ \hline
        ESP EM($\leq40$) & \textbf{7.4\%} & \textbf{15.5\%} & 29.8\% & 41.0\% \\
        ESP EM($\leq80$) & 7.1\% & 14.7\% & \textbf{29.9\%} & \textbf{44.8\%} \\
        ESP Recall($\leq40$) & \textbf{37.3\%} & \textbf{46.1\%} & 58.4\% &66.4\% \\ 
        ESP Recall($\leq80$) & 34.9\% & 44.6\% & \textbf{59.1\%} & \textbf{70.0\%} \\\Xhline{1pt}
    \end{tabular}
    \caption{Evaluation of reasoning chains by ESP scores on two versions of the entity graphs in the development set. $\leq40$ and $\leq80$ indicate to the maximum number of nodes in entity graphs. Note that $\leq40$ refers to the entity graph whose entities are extracted by Stanford CoreNLP, while $\leq80$ refers to the entity graph whose entities are extracted by the aforementioned BERT NER model.}
    \label{tab:esp}
\end{table}

We train a DFGN with 2 fusion blocks to select paths with top-$k$ scores. In the development set, the average number of paths of length 2 is 174.7. We choose $k$ as $1,2,5,10$ to compute ESP EM and ESP Recall scores. As we can see in Table \ref{tab:esp}, regarding the supporting sentences as the ground truth of reasoning chains, our framework can predict reliable information flow. The most informative flow can cover the supporting facts and help produce reliable reasoning results.
Here we present the results from two versions of the entity graphs. 
The results with a maximum number of nodes $\leq40$ are from the entity graph whose entities are extracted by Stanford CoreNLP.
The results with a maximum number of nodes $\leq80$ are from the entity graph whose entities are extracted by the aforementioned BERT NER model.
Since the BERT NER model performs better, we use a larger maximum number of nodes.

In addition, as the size of an entity graph gets larger, the expansion of reasoning chain space makes a Hit even more difficult. However, the BERT NER model still keeps comparative and even better performance on metrics of EM and Recall.
Thus the entity graph built from the BERT NER model is better than the previous version.


\begin{figure*}[t]
\begin{center}
\includegraphics[width=2.05\columnwidth]{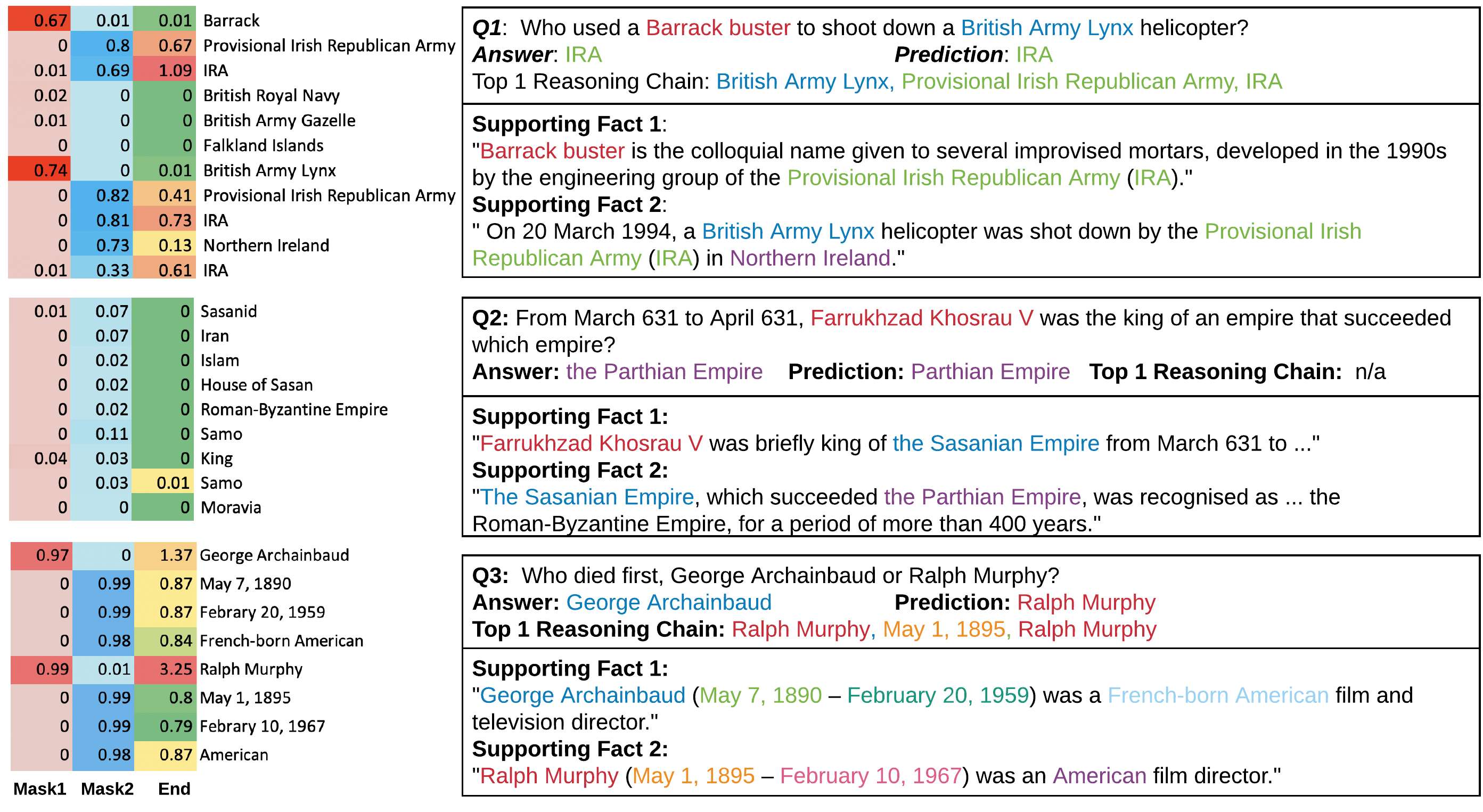}
\end{center}
\caption{Case study of three samples in the development set. We train a DFGN with 2-layer fusion blocks to produce the results. The numbers on the left side indicate the importance scores of the predicted masks. The text on the right side include the queries, answers, predictions, predicted top-1 reasoning chains and the supporting facts of three samples with the recognized entities highlighted by different colors.}
\label{fig:casestudy}
\end{figure*}

\subsection{Case Study}
We present a case study in Figure \ref{fig:casestudy}.
The first case illustrates the reasoning process in a DFGN with 2-layer fusion blocks. At the first step, by comparing the query with entities, our model generates \textbf{Mask1} as the start entity mask of reasoning, where \textit{``Barrack''} and \textit{``British Army Lynx''} are detected as the start entities of two reasoning chains. Information of two start entities is then passed to their neighbors on the entity graph. At the second step, mentions of the same entity \textit{``IRA''} are detected by \textbf{Mask2}, serving as a bridge for propagating information across two paragraphs. Finally, two reasoning chains are linked together by the bridge entity \textit{``IRA''}, which is exactly the answer.

The second case in Figure \ref{fig:casestudy} is a bad case. Due to the malfunction of the NER module, the only start entity, \textit{``Farrukhzad Khosrau V''}, was not successfully detected. Without the start entities, the reasoning chains cannot be established, and the further information flow in the entity graph is blocked at the first step.

The third case in Figure \ref{fig:casestudy} is also a bad case, which includes a query of the \textit{Comparison} query type. Due to the lack of numerical computation ability of our model, it fails to give a correct answer, although the query is just a simple comparison between two days \textit{``February 20, 1959''} and \textit{``February 10, 1967''}. It is an essential problem to incorporate numerical operations for further improving the performance in cases of the comparison query type.

\section{Conclusion}
\label{sec:conclusion}
We introduce Dynamically Fused Graph Network (\method) to address multi-hop reasoning. Specifically, we propose a dynamic fusion reasoning block based on graph neural networks. Different from previous approaches in QA, \method is capable of predicting the sub-graphs dynamically at each reasoning step, and the entity-level reasoning is fused with token-level contexts. We evaluate \method on \hotpotqa and achieve leading results. Besides, our analysis shows \method can produce reliable and explainable reasoning chains. In the future, we may incorporate new advances in building entity graphs from texts, and solve more difficult reasoning problems, e.g. the cases of comparison query type in \hotpotqa.

\bibliography{acl2019}
\bibliographystyle{acl_natbib}
\end{document}